\documentclass[10pt,twocolumn,letterpaper]{article}

\usepackage[pagenumbers]{cvpr} 

\definecolor{cvprblue}{rgb}{0.21,0.49,0.74}
\usepackage[pagebackref,breaklinks,colorlinks,allcolors=cvprblue]{hyperref}
\usepackage{url}            
\usepackage{booktabs,makecell,fvextra}       
\usepackage{amsfonts}       
\usepackage{nicefrac}       
\usepackage{microtype}      
\usepackage{xcolor}         
\usepackage{amsmath,amssymb,graphicx,amsthm}
\usepackage{enumitem,listings}
\usepackage{inconsolata}
\usepackage[T1]{fontenc}
\usepackage{orcidlink,textcomp}
\usepackage{float,siunitx,caption}
\usepackage{multirow,diagbox,calc,cuted}
\usepackage{subcaption}
\usepackage{stfloats}
\usepackage{needspace}

\usepackage[textsize=tiny]{todonotes}

\def\x{{\mathbf x}}
\def\X{{\mathcal X}}
\def\xhat{{\mathbf{\hat x}}}
\def\y{{\mathbf y}}
\def\A{{\mathbf A}}

\def\M{{\mathbf M}}
\def\T{{\mathbf T}}
\def\Tg{{\mathbf{T}_g}}

\def\L{{\mathcal{L}}}
\def\GT{{\text{GT}}}
\def\ft{{f_\theta}}

\definecolor{codegreen}{rgb}{0,0.6,0}
\definecolor{codegray}{rgb}{0.5,0.5,0.5}
\definecolor{codepurple}{rgb}{0.58,0,0.82}
\definecolor{backcolour}{rgb}{0.95,0.95,0.92}
\lstdefinestyle{mystyle}{
    commentstyle=\color{codegreen},
    keywordstyle=\color{magenta},
    numberstyle=\tiny\color{codegray},
    stringstyle=\color{codepurple},
    basicstyle=\ttfamily\scriptsize,
}
\theoremstyle{definition}

\def\paperID{8} 
\def\confName{CVPR}
\def\confYear{2026}

\title{Perspective-Equivariant Fine-tuning for Multispectral Demosaicing\\ without Ground Truth}

\author{
  Andrew Wang \orcidlink{0000-0003-0838-7986}\\
  School of Engineering\\
  University of Edinburgh\\
  \and
  Mike Davies \orcidlink{0000-0003-2327-236X}\\
  School of Engineering\\
  University of Edinburgh\\
}

\begin{document}
\maketitle
\begin{abstract}
Multispectral demosaicing is crucial to reconstruct full-resolution spectral images from snapshot mosaiced measurements, enabling real-time imaging from neurosurgery to autonomous driving. Classical methods are blurry, while supervised learning requires costly ground truth (GT) obtained from slow line-scanning systems. We propose \textbf{P}erspective-\textbf{E}quivariant \textbf{F}ine-tuning for \textbf{D}emosaicing (\textbf{PEFD}), a framework that learns multispectral demosaicing from mosaiced measurements alone. PEFD a) exploits the projective geometry of camera-based imaging systems to leverage a richer group structure than previous demosaicing methods to recover more null-space information, and b) learns efficiently without GT by adapting pretrained foundation models designed for 1-3 channel imaging. On surgical and automotive datasets, PEFD recovers fine details such as blood vessels and preserves spectral fidelity, substantially outperforming recent approaches, nearing supervised performance. Furthermore, the performance of PEFD is demonstrated on raw, unprocessed data from a commercial multispectral sensor. Code is at \url{https://github.com/Andrewwango/pefd}.
\end{abstract}    
\section{Introduction}
\label{sec:intro}

Multispectral imaging (MSI) captures spectral information across multiple wavelengths, providing a non-contact, non-ionising solution that reveals diagnostic information beyond conventional RGB imaging. In intraoperative neurosurgery or oral cancer imaging, MSI exploits the inherent optical properties of tissue to discriminate between tumour and healthy tissue, overcoming limitations of subjective visual assessment \citep{clancy_surgical_2020,chand_-vivo_2024}, or provides quantitative tissue perfusion and oxygen saturation measurements across a wide field of view \citep{holmer_hyperspectral_2018}. In oral surgery, MSI enables early detection of oral lesions \citep{hyttinen_oral_2020}. In autonomous driving, MSI enables robust object detection and scene understanding under varying illumination and weather conditions where RGB cameras fail. In remote sensing, MSI distinguishes classes that appear identical in RGB imagery for vegetation monitoring. Depending on the number of bands, the technology may be referred to as multispectral or hyperspectral; the methods in this work apply to both.

Snapshot multispectral cameras employ multispectral filter arrays (MSFAs) to acquire spectral data in a single exposure, offering significant speed and cost advantages over traditional line-scanning systems, yet do not require complex optics seen in compressive systems such as CASSI \citep{bacca_computational_2023}. By capturing all spectral bands simultaneously, snapshot imaging via mosaicing \citep{li_deep_2022} eliminates motion artifacts, enables real-time acquisition, and uses compact hardware suitable for integration into surgical workflows, vehicle-mounted systems, and UAV platforms\citep{huang_spectral_2022}. However, this comes at the cost of reduced spatial and spectral resolution, as each pixel captures only one spectral band. Demosaicing algorithms are therefore required to reconstruct the full-resolution multispectral image from the mosaiced measurements.

\begin{figure}[tb]
  \centering
  \includegraphics[width=0.5\textwidth]{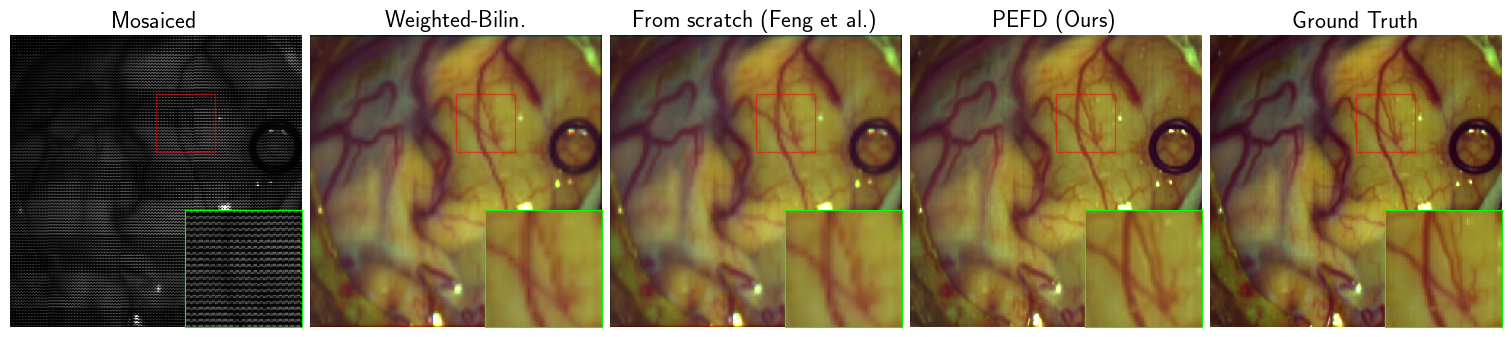}
  \caption{Perspective-Equivariant Fine-tuning for Demosaicing (PEFD) recovers sharp multispectral images with correct colours without ground truth, compared to popular classical methods such as weighted bilinear demosaicing or learning from scratch.}
  \label{fig:results_abridged}
\end{figure}

Specifically, let $\y \in \mathbb{R}^{H \times W \times 1}$ be the mosaiced sensor measurement, where each spatial location captures a single spectral band determined by the MSFA pattern. Let $\x \in \mathbb{R}^{H \times W \times C}$ be the unknown ground truth (GT) multispectral image with $C$ spectral channels. These are then related by

\begin{equation}
    \y = \A\x + \mathbf{\epsilon}
    \label{eq:forward}
\end{equation}

\noindent where $\A$ is the mosaicing operator that selects one spectral channel per spatial location according to the MSFA pattern of shape $c\times c=C$, and $\mathbf{\epsilon}$ is additive noise. While many different design choices are possible for the MSFA \citep{lapray_multispectral_2014}, the methods considered here are agnostic to the specific pattern and generalise to arbitrary patterns. \cref{eq:forward} constitutes a highly ill-posed inverse problem, as the number of measurements $m=HW\ll n=HWC$, the total number of spatial-spectral samples.

Classical demosaicing methods employ interpolation-based techniques such as bilinear or Gaussian, or statistics-based approaches \citep{brauers_color_2006,mihoubi_multispectral_2017,yu_colour_2006}. While computationally efficient, these methods suffer from spectral artifacts and spatial blurring, particularly for fine structures. Other methods use variational optimisation algorithms from compressed sensing, employing handcrafted regularisers e.g. total variation (TV) \citep{chambolle_introduction_2016}, but these are limited by the need to craft complex priors for individual problems, lengthy test-time optimisation, and poor performance for highly sparse measurements.

Supervised deep learning circumvents the pitfalls of previous approaches and has demonstrated superior reconstruction quality for demosaicing tasks. These methods train a neural network $\ft$ to reconstruct $\xhat = \ft(\y)$ using a loss $\L_\text{sup} = \L(\ft(\y), \x_\GT)$ where $\x_\GT$ is ground truth (GT), and many such networks have been proposed \citep{arad_ntire_2022,habtegebrial_deep_2019,zhou_pidsr_2025}. 

However, supervised methods require large datasets of high-resolution paired GT images. Therefore, these methods are useless when GT images are prohibitively expensive or even impossible to obtain in many practical scientific, medical or environmental use-cases, resulting in a chicken-and-egg problem: how can a system train with high quality images, if these images cannot be acquired? In particular, the acquisition of high-quality, pixel-aligned multispectral GT image acquisition typically requires slow and/or bulky line-scanning systems \citep{fabelo_-vivo_2019,hyttinen_oral_2020,winkens_hyko_2017} incompatible with real-time applications or hardware constraints, which are typical in such applications. This creates a fundamental limitation: supervised models cannot adapt to new domains without costly GT acquisition, and may never learn to reconstruct images of phenomena that have never been fully observed.

Self-supervised methods offer a solution by training a model $\ft$ to demosaic from mosaiced measurements alone, acquired from a snapshot sensor without knowledge of GT \cite{feng_unsupervised_2024,garcia-barajas_self-supervised_2025}. We note that Deep Image Prior approaches \citep{ulyanov_deep_2018,li_joint_2024} optimise network parameters individually for each image using measurement consistency. However, this requires lengthy test-time optimisation, unsuitable for real-time applications.

While fully self-supervised methods are conceptually important, in practice they generally train networks from scratch, and thus a) result in subpar performance in limited data scenarios, and b) do not benefit from the vast swathes of high quality related GT data that have been painstakingly acquired in recent years \citep{fabelo_-vivo_2019,hyttinen_oral_2020,winkens_hyko_2017,li_lsdir_2023,agustsson_ntire_2017}. These large-scale datasets have led to the emergence of foundation image restoration models that are robust across several imaging tasks and modalities \cite{terris_reconstruct_2025}. However, these pretrained models inevitably do not offer competitive performance out-of-the-box on specific modalities such as demosaicing, and benefit from additional self-supervised fine-tuning. In this paper, we present \textbf{P}erspective-\textbf{E}quivariant \textbf{F}ine-tuning for \textbf{D}emosaicing (PEFD), a demosaicing framework that combines the efficient adaptation to new domains offered by self-supervised losses with the robustness of pretrained image restoration models. We propose a GT-free loss for demosaicing that leverages the projective geometry of optical camera-based imaging systems. Our \textbf{contributions} are:

\begin{enumerate}
    \item A self-supervised loss for multispectral demosaicing, that exploits perspective-equivariance of natural images;
    \item A framework for fine-tuning image restoration networks to demosaic without GT nor large-scale training data;
    \item Extensive experimental validation and comparisons on surgical and automotive applications, demonstrating state-of-the-art unsupervised demosaicing performance.
\end{enumerate}
\section{Related Work}
\label{sec:related}

\subsection{Self-supervised demosaicing}

Recent self-supervised demosaicing methods have demonstrated promising results on natural scenes. \citet{feng_unsupervised_2024} train SDNet, a spectral-spatial attention network, using the shift-equivariant loss from \citet{chen_equivariant_2021} on controlled test scenes. \citet{garcia-barajas_self-supervised_2025} adapt DnCNN \citep{zhang_beyond_2017}, a network originally proposed for denoising, for multispectral channels and employ the rotation-equivariant loss from \cite{chen_robust_2022}. Noise2Noise-style methods \citep{ehret_joint_2019} split the measurements and train by using one measurement at input and another as target, co-acquired \eg during a burst; however, this requires multiple acquisitions which is unrealistic for many cameras in practice. \citet{li_spatial_2023} apply measurement consistency (MC) combined with classical regularisation terms (total variation, Tikhonov, and inter-channel correlation) to brain intraoperative imaging \cite{fabelo_-vivo_2019}. However, MC cannot recover information in the null-space $\mathcal{N}(\A)$ of the mosaicing operator, as any solution $\A^\dagger\y+\mathbf{v}$ trivially satisfies $\L_\text{MC}=0$ where $\mathbf{v}\in\mathcal{N}(\A)$; this approach thus reduces to classical regularisation, where terms such as TV are selective in the null-space and promotes a predefined simplistic sparsity-based image model. \citet{li_self-supervised_2024} extend this work with a cyclic adversarial loss. However, this requires a full-resolution RGB dataset to learn the spatial details lost in the null-space, as well as training of highly unstable generative adversarial networks \cite{goodfellow_generative_2014}, which is beyond our scope.

Beyond demosaicing, self-supervised methods have been developed for various imaging tasks including MRI reconstruction \cite{wang_benchmarking_2025,millard_clean_2024}, denoising \cite{tachella_unsure_2025,monroy_generalized_2025,huang_neighbor2neighbor_2021,krull_noise2void_2019}, and pan-sharpening \cite{ciotola_unsupervised_2023,wang_perspective-equivariance_2024}. A powerful framework that emerges from these works is Equivariant Imaging (EI) \cite{chen_equivariant_2021}, which constrains the solution space using group invariance, for example to shifts \citep{chen_equivariant_2021} for tomography, rotations \cite{chen_robust_2022} for MRI, or scale \citep{scanvic_self-supervised_2023} for deblurring. However, none of these EI methods have been applied to demosaicing, which is a particularly challenging problem compared to previous applications of EI, due to the high sampling sparsity with no global information provided by the physics (unlike \eg pan-sharpening \citep{wang_perspective-equivariance_2024}, where both spectral and high-frequency spatial guidance is provided). Additionally, the MSFA pattern results in periodic subsampling, which cannot be handled well with shift or rotation \citep{scanvic_self-supervised_2023}; observe that a general mosaicing operator is equivariant to both $c$-pixel shifts $\textbf{C}_{\frac{W}{c}}$ and rotations $\textbf{C}_4$. Finally, they all train models from scratch without leveraging pretrained weights.

Deep Image Prior (DIP) methods \citep{ulyanov_deep_2018,darestani_accelerated_2021,li_unsupervised_2022,kurniawan_noise-resistant_2022,park_joint_2020,li_joint_2024} and PnP approaches \cite{zhang_plug-and-play_2022,lai_deep_2022} perform iterative optimisation at test time. However, these require lengthy inference unsuitable for real-time applications. Furthermore, PnP methods still rely on well-trained denoisers, merely shifting the GT requirement from the reconstructor to the denoiser, which also would benefit from fine-tuning as discussed in \cref{sec:related_ft} in order to remain competitive.

We note that several works address joint demosaicing and denoising (JDD) \citep{li_joint_2024,park_joint_2020,garcia-barajas_self-supervised_2025}. However, since denoising is a supplementary problem to that of recovering information from the null-space of the mosaicing operator, we note that, following established practice \citep{millard_clean_2024,chen_robust_2022}, any self-supervised demosaicing method discussed here can be extended to JDD by appending a self-supervised denoising loss \citep{stein_estimation_1981,monroy_generalized_2025}.

\subsection{Fine-tuning without ground truth}
\label{sec:related_ft}

While self-supervised methods enable learning without GT for the specific domain and hardware, training from scratch discards the knowledge encoded in models that have been trained on vast amounts of related high-quality data across multiple adjacent imaging tasks and modalities \cite{terris_reconstruct_2025}. Furthermore, their performance is likely to be subpar in limited data scenarios, which are common in scientific imaging. These pretrained models offer strong inductive bias but require adaptation to specific domains, modalities or particular hardware where they do not perform competitively out-of-the-box. Recent work has explored fine-tuning pretrained models without GT. In MRI, \citet{darestani_test-time_2022} fine-tune using measurement consistency, but as discussed above, this cannot recover null-space information in highly sparse settings. \citet{liu_fine-tuning_2024} propose fine-tuning with within-filter shift-equivariance (\ie pixel-shifts of $<\pm c$) for RGB demosaicing on natural images; this is a small group $\textbf{C}_C$, and thus has limited ability to sufficiently regularise the problem. Furthermore, the method is limited to 3-channel data, and only applied to basic pretrained models such as UNet \citep{ronneberger_u-net_2015}. Reconstruct Anything Model \citep{terris_reconstruct_2025} is a foundation model trained across grayscale, complex, and RGB data on tasks including inpainting, deblurring, denoising, super-resolution, and medical imaging reconstruction, and demonstrated fine-tuning capability for RGB demosaicing, though reconstructions lacked sharpness and were limited to 3 channels. In contrast, our work proposes self-supervised learning via perspective-equivariance, to which the mosaicing operator is not equivariant, to efficiently adapt a robust pretrained model to multispectral demosaicing. 
\section{Methodology}
\label{sec:method}

\subsection{Projective geometry in camera systems}

\begin{figure}[tb]
  \centering
  \includegraphics[width=\linewidth]{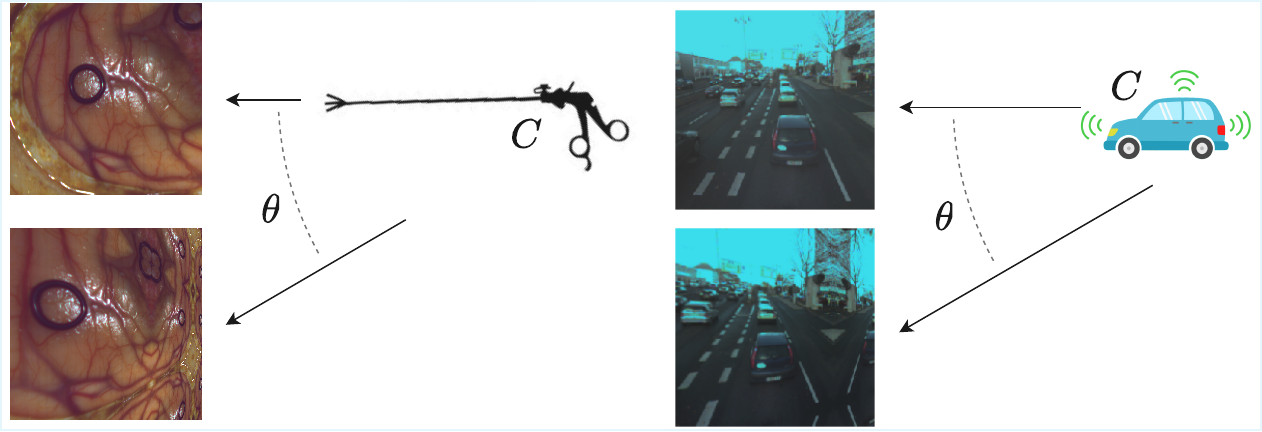}
  \caption{Given a camera centre $C$, camera systems in intraoperative (left) and automotive (right) imaging rotate freely about their axes, producing images related by perspective transformations.}
  \label{fig:camera_systems}
\end{figure}

Multispectral imaging systems are often mounted on platforms that move and rotate freely in the world. Intraoperative cameras navigate the surgical field during procedures, while automotive multispectral cameras mounted on vehicles observe driving scenes from continuously varying viewpoints. In both cases, the camera orientation changes relative to the observed scene, producing images from different perspectives; see \cref{fig:camera_systems} for an example. Following projective geometry \cite{hartley_multiple_2004}, images captured by a camera rotating about its centre are related by projective transformations (homographies). Specifically, let the homogeneous coordinates of two images $\x,\x^\prime\in\mathbb{R}^n$ of the same scene taken from different camera orientations be $\chi,\chi^\prime\in\mathbb{R}^{3\times n}$; the transform between them $\x^\prime=\Tg(\x)$ can then be writtten linearly as $\chi^\prime=\Upsilon_g\chi$, where $\Upsilon_g\in\mathbb{R}^{3\times 3}$ and $g\in G$ is drawn from the group of homographies. This transformation decomposes as:

\begin{equation}
\Upsilon_g=\mathbf{K}\mathbf{R}_z(\theta_z)\mathbf{R}_y(\theta_y)\mathbf{R}_x(\theta_x)\mathbf{K}^{-1}
\label{eq:perspective_transform}
\end{equation}

\noindent where $\mathbf{K}$ contains camera intrinsics (focal length, principal point, pixel dimensions), and $\mathbf{R}_x,\mathbf{R}_y,\mathbf{R}_z$ are rotation matrices about the camera axes parametrised by Euler angles $\theta_x,\theta_y,\theta_z$. The pan (rotation about $x$-axis, $\theta_x$) and tilt (s$\theta_y$) transformations create the characteristic perspective effect where parallel lines converge. This perspective transformation group contains previously studied groups as special cases \eg rotation, shifts or scale \citep{chen_equivariant_2021,chen_robust_2022,scanvic_self-supervised_2023}. Note that the group does not induce other distortions (\eg stretching) if the intrinsics are kept constant. The decomposition \cref{eq:perspective_transform} provides an easy way to randomly sample physically plausible transforms by sampling $\theta_x,\theta_y,\theta_z$.

We posit that the unknown set of multispectral images $\X$ is invariant to these perspective transformations, i.e., $\mathbf{T}_g(\x)\in\X$ for all $g\in G,\x\in\X$. This assumption is natural for camera-based imaging: an image of a surgical field or driving scene remains a valid image of that environment when viewed from a slightly different angle, as the surgeon or car rotates freely in the world; see examples in \cref{fig:camera_systems}. Note that the transforms do not reveal any occluded information; they merely relate images taken from the same camera at different angles. Crucially, this invariance holds even though we lack access to GT images (\ie data augmentation is impossible).

\subsection{Perspective-equivariant fine-tuning for demosaicing}

\begin{figure}[tb]
  \centering
  \includegraphics[width=\linewidth]{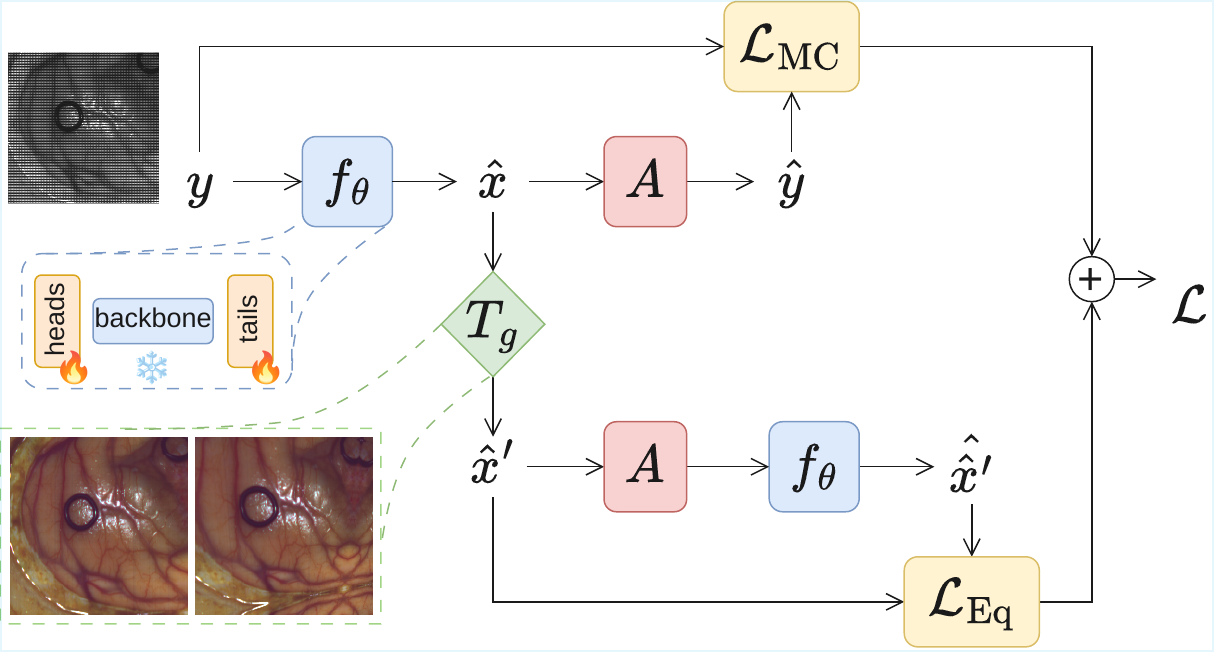}
  \caption{Framework for perspective-equivariance fine-tuning for multispectral demosaicing, where $\ft$ is our adapted pretrained foundation model \citep{terris_reconstruct_2025} to process multispectral data, $\y$ are the mosaiced measurements, $\A$ is the mosaicing operator, $\Tg$ is the parametrised perspective transformation and the loss depicts \cref{eq:ei_loss}.}
  \label{fig:framework}
\end{figure}

The Equivariant Imaging (EI) framework \cite{chen_equivariant_2021} exploits group invariance to recover information beyond what measurement consistency alone can provide. Given mosaiced measurements $\y=\M\x$, the measurement consistency loss $\L_\text{MC}=\lVert\A\ft(\y)-\y\rVert_2^2$ cannot recover any component in the null-space $\mathcal{N}(\A)$, as any reconstruction $\A^\dagger\y+\mathbf{v}$ with $\mathbf{v}\in\mathcal{N}(\A)$ trivially satisfies $\L_\text{MC}=0$. EI overcomes this limitation by observing that if $\X$ is $G$-invariant, then $\y=\A\x=\A\circ\T_g^{-1}\circ\mathbf{T}_g(\x)=\A_g\x^\prime$ where $\A_g=\A\circ\mathbf{T}_g^{-1}$ is a virtual forward operator and $\x^\prime=\mathbf{T}_g(\x)\in\X$. By sampling transformations $g\in G$, we obtain a family of virtual operators $\{\A_g\}$ that collectively contain information about the null-space. The EI loss enforces consistency between reconstructions under these transformations:

\begin{equation}
\underbrace{\lVert\A\ft(\y)-\y\rVert_2^2}_{\text{MC}} + \alpha\underbrace{\lVert \mathbf{T}_g(\ft(\y))-\ft(\A\circ\mathbf{T}_g(\ft(\y))) \rVert_2^2}_{\text{equivariance}}
\label{eq:ei_loss}
\end{equation}

\noindent where $\alpha=0.1$ is a weighting parameter, and $g\in G$ is sampled during training. The loss is depicted in \cref{fig:framework}. The perspective transformation group $G$, studied originally for satellite pan-sharpening \citep{wang_perspective-equivariance_2024}, offers substantially richer structure than previously studied groups in demosaicing: pixel shifts \cite{feng_unsupervised_2024} and $\text{SO}(2)$ \cite{garcia-barajas_self-supervised_2025} are strict subgroups of $G$, meaning we can exploit more symmetries and probe a larger set of virtual operators, to recover details lost in mosaicing. We refer the reader to \citet{wang_perspective-equivariance_2024} for theoretical details. Shifts and rotations moreover cannot recover high-frequency information from the null-space \citep{scanvic_self-supervised_2023}. The parametrisation in \cref{eq:perspective_transform} also allows us to sample physically plausible transformations during training.

While previous methods \citep{feng_unsupervised_2024,garcia-barajas_self-supervised_2025} trained models from scratch, we posit that instead fine-tuning robust pretrained models leads to improved demosaicing, especially in limited-data scenarios. Therefore, we combine the perspective-equivariant loss with fine-tuning of a pretrained foundation model. Reconstruct Anything Model (RAM) \cite{terris_reconstruct_2025} was trained on diverse imaging tasks including inpainting, deblurring, denoising, super-resolution, and medical image reconstruction across grayscale, complex, and RGB data. While RAM demonstrates strong and robust performance across these tasks, it was not trained nor fine-tuned beyond three-channel data and therefore exhibits poor out-of-the-box performance on multispectral demosaicing.

We adapt the 1-3 channel RAM architecture to fine-tune for multispectral demosaicing in a parameter-efficient manner, by freezing the 32M-parameter convolutional encoder-decoder backbone and replicating the grayscale channel-specific heads and tail for multispectral $C$ channels. This strategy leverages the robust feature representations learned by the backbone while adapting the task-specific layers to multispectral demosaicing, as well as preventing overfitting to the self-supervised loss, and enabling efficient fine-tuning with limited samples.

We refer to this complete framework in \cref{fig:framework} as \textbf{P}erspective-\textbf{E}quivariant \textbf{F}ine-tuning for \textbf{D}emosaicing (PEFD). By combining perspective-equivariance with foundation model fine-tuning, PEFD exploits both the knowledge encoded in pretrained weights and the null-space recovery enabled by the richer group structure, achieving high-quality multispectral reconstruction without GT.

\subsection{Extension to joint demosaicing and denoising}

PEFD naturally extends to joint demosaicing and denoising (JDD) in low-light scenarios where shot noise is significant, following established practice \citep{garcia-barajas_self-supervised_2025,li_joint_2024}, by augmenting the loss with a GT-free denoising term \citep{monroy_generalized_2025,huang_neighbor2neighbor_2021}. This loss can be chosen depending on how much information about the noise model is known \citep{tachella_unsure_2025}. Since denoising is a supplementary problem to that of recovering information from the null-space of the mosaicing operator, we leave evaluation of this extension to future work.
\section{Experiments \& Analysis}
\label{sec:experiments}

We demonstrate the performance of our proposed PEFD method on two real imaging datasets covering medical and automotive scenarios, by implementing the framework with DeepInverse \citep{tachella_deepinverse_2025}, and compare against classical interpolation methods, optimisation-based approaches, as well as recent self-supervised demosaicing methods. Code is available at \url{https://github.com/Andrewwango/pefd}.

\subsection{Experimental setup}

\begin{figure*}[tb]
  \centering
  \includegraphics[width=0.99\textwidth]{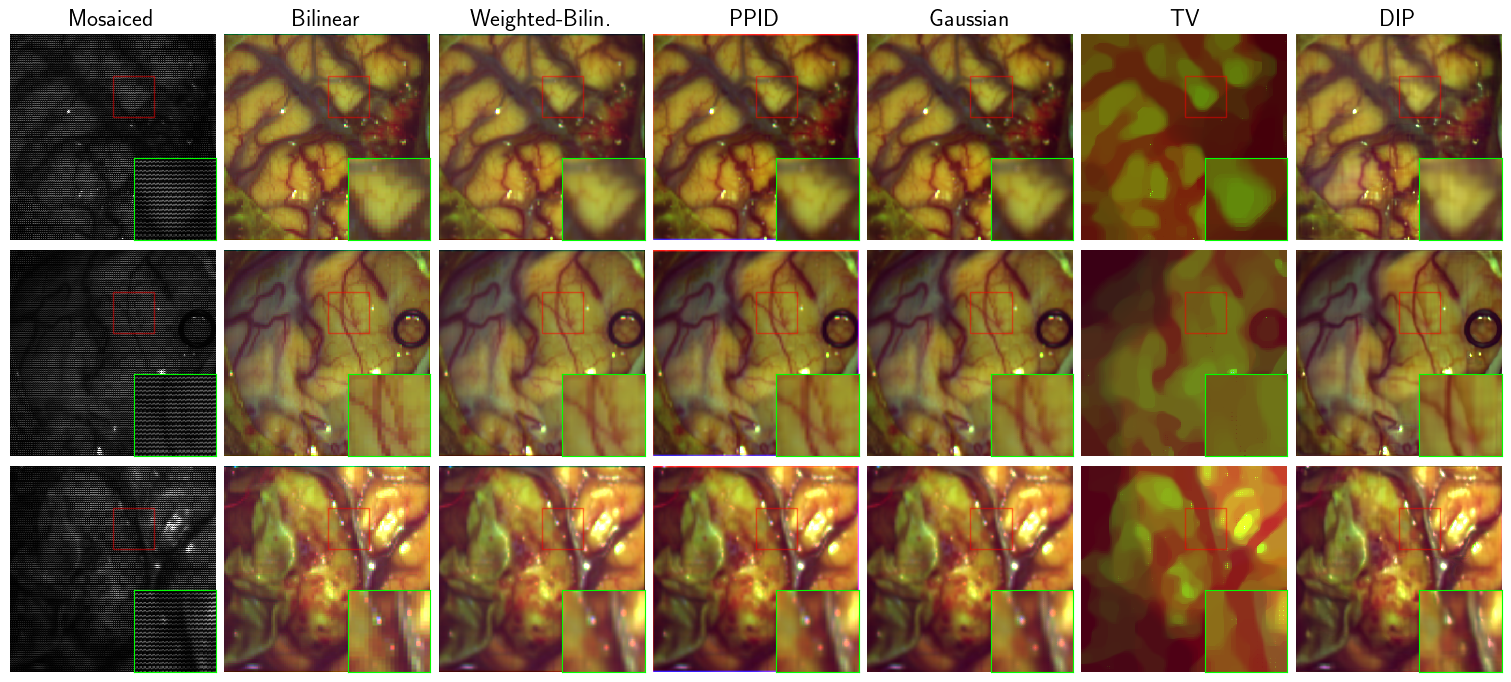}
  \includegraphics[width=0.99\textwidth]{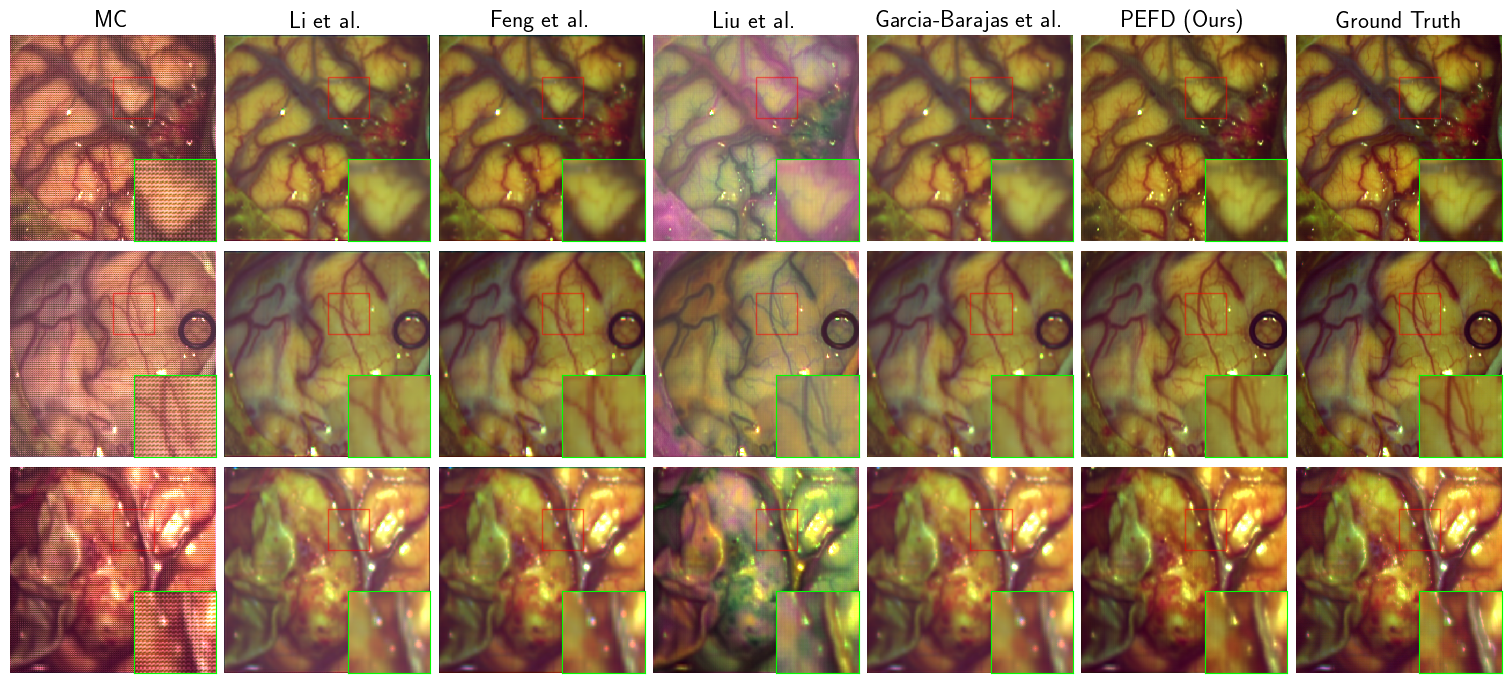}
  \caption{Test set false-RGB reconstruction results on 3 example neurosurgical intraoperative images from HELICoiD \cite{fabelo_-vivo_2019}, showing mosaiced input $\y$, reconstruction results for 9 comparison methods, our reconstruction with PEFD, and ground truth for reference.}
  \label{fig:results_helicoid}
\end{figure*}

\subsubsection{Data}
We employ two publicly available datasets of real-world spectral images captured for neurosurgical and automative use-cases. Furthermore, we consider the realistic scenario of a limited training set of mosaiced measurements. The HELICoiD dataset \citep{fabelo_-vivo_2019,leon_hyperspectral_2023} comprises in total 70, split 80\%/20\%, in-vivo hyperspectral images of human brain tissue, acquired using a Hyperspec VNIR A-Series pushbroom camera covering 400--1000~nm during neurosurgical operations. Following \citet{li_deep_2022}, we simulate 16 multispectral bands from the original hyperspectral data and crop to $384\times384$. We generate mosaiced measurements using the $4\times4$ sequential MSFA pattern, which is widely used in practice \citep{ximec_xispec2_2021}, though our method does not leverage a specific pattern and therefore can be used with arbitrary MSFAs. Secondly, the HyKo dataset \citep{winkens_hyko_2017} contains images captured with an imec-Ximea MQ022HG camera (470--630~nm) mounted on land-based vehicles under various driving and lighting conditions. Again, mosaiced data is simulated with a $4\times4$ MSFA for 70 volumes, and cropped to $248\times248$.

Finally, we conduct a raw data experiment to validate our method on real, unprocessed snapshot sensor data, where GT is truly unavailable. We fine-tune and test our method on unprocessed data from the imec CMV2K-SSM4x4-VIS CMOS sensor (460--600~nm) captured during oral cancer imaging from \cite{chand_-vivo_2024}, before any image corrections are made. The imec sensor uses a $1024\times 1088$ pixel array with the same sequential $4\times 4$ MSFA \citep{ximec_xispec2_2021}, leading to multispectral images of shape $512\times 272$. Thus, the goal of demosaicing is to recover the multispectral image at full $1024\times 1088$ resolution.

\subsubsection{Model}
We adapt the pretrained foundation model RAM \citep{terris_reconstruct_2025}, designed for 1-3 channel data, to perform multispectral demosaicing by freezing the backbone and replicating the grayscale weights across 16 channels to initialise the heads and tails. The base model is labelled ``RAM zero-shot'' in the ablation. All compared models receive as input the Gaussian interpolated image as an initial reconstruction, and are trained for 200 epochs with learning rate 1e-5, on a NVIDIA H100 GPU.

\begin{figure*}[tb]
  \centering
  \vspace{1em} 
  \begin{subfigure}[b]{0.5\textwidth}
    \centering
    \includegraphics[width=\textwidth]{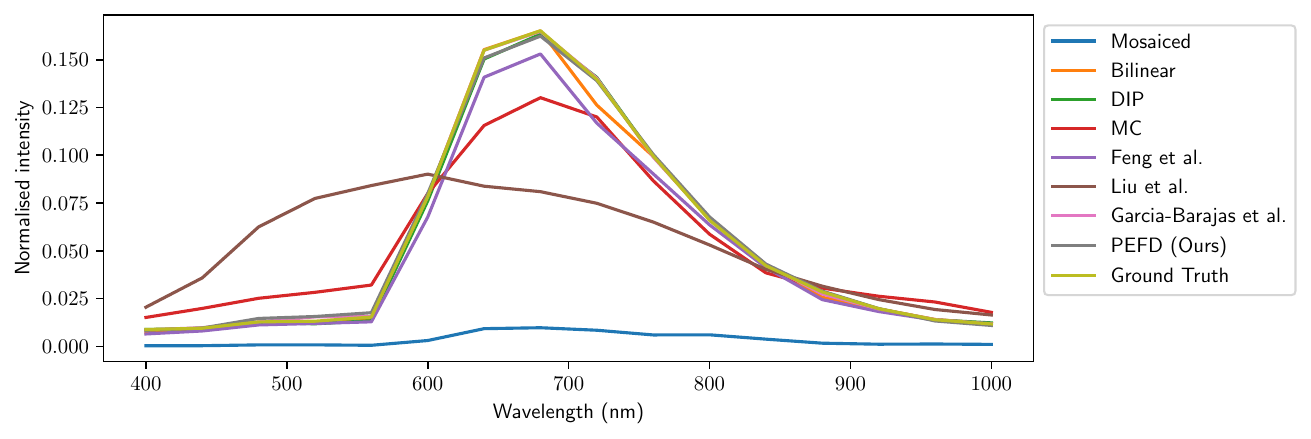}
  \end{subfigure}
  \begin{subfigure}[b]{0.4\textwidth}
    \centering
    \includegraphics[width=\textwidth]{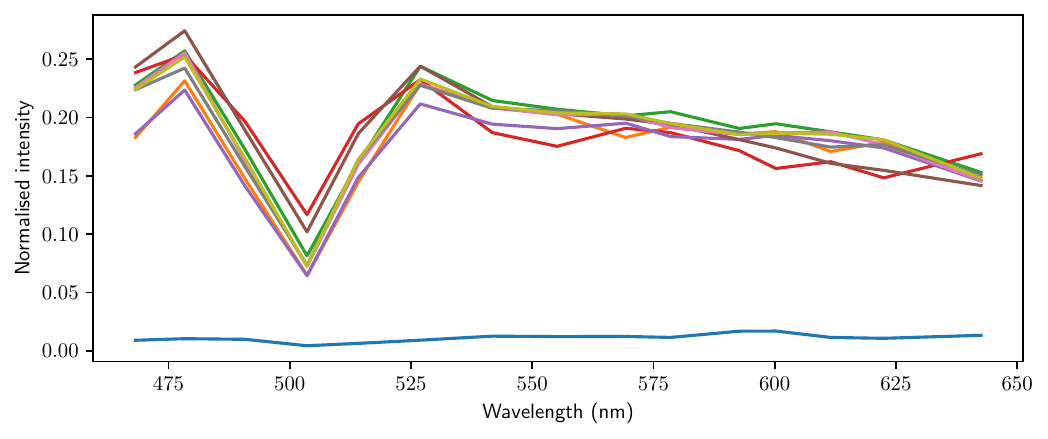}
  \end{subfigure}
  \caption{Example spectral signatures from a sample image patch in each dataset (left: HELICoiD; right: HyKo).}
  \label{fig:spectral_examples}
\end{figure*}

\begin{figure*}[tb]
\centering
  \includegraphics[width=0.999\textwidth]{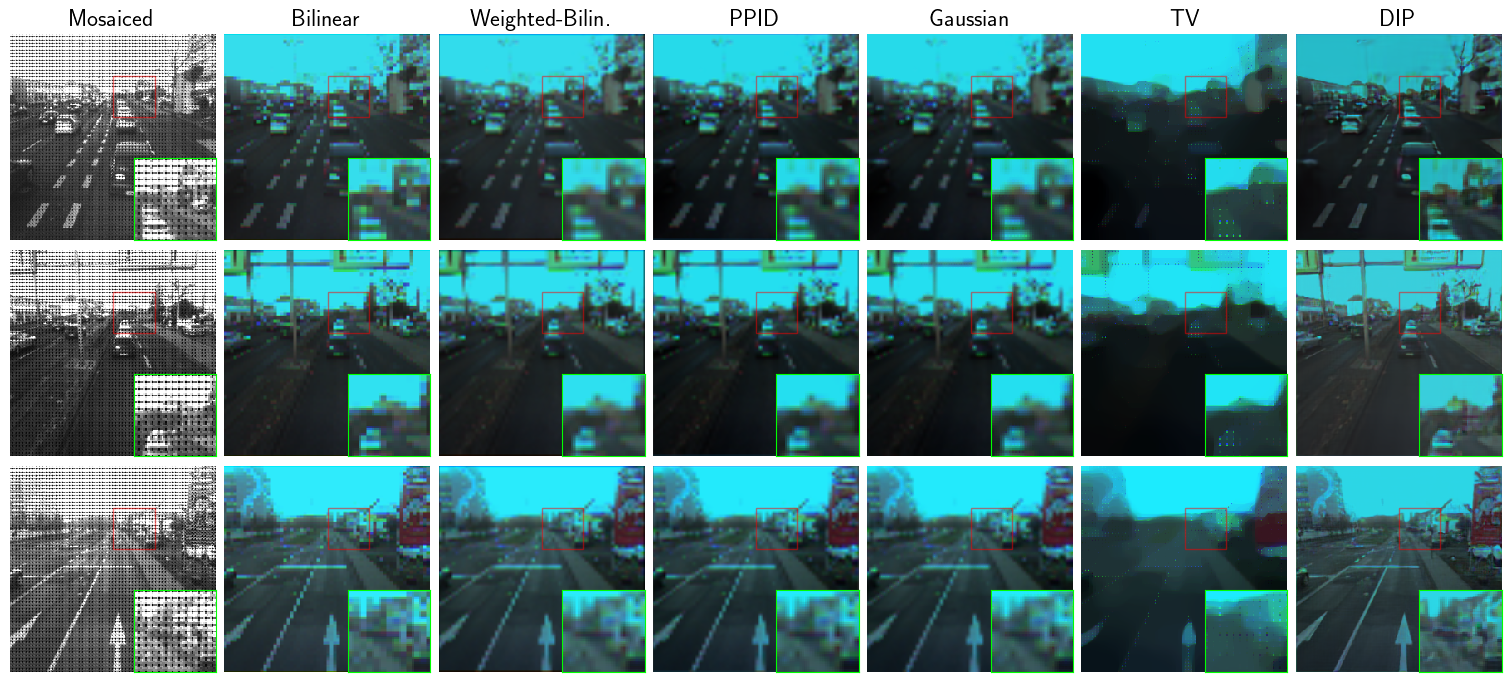}
  \includegraphics[width=0.999\textwidth]{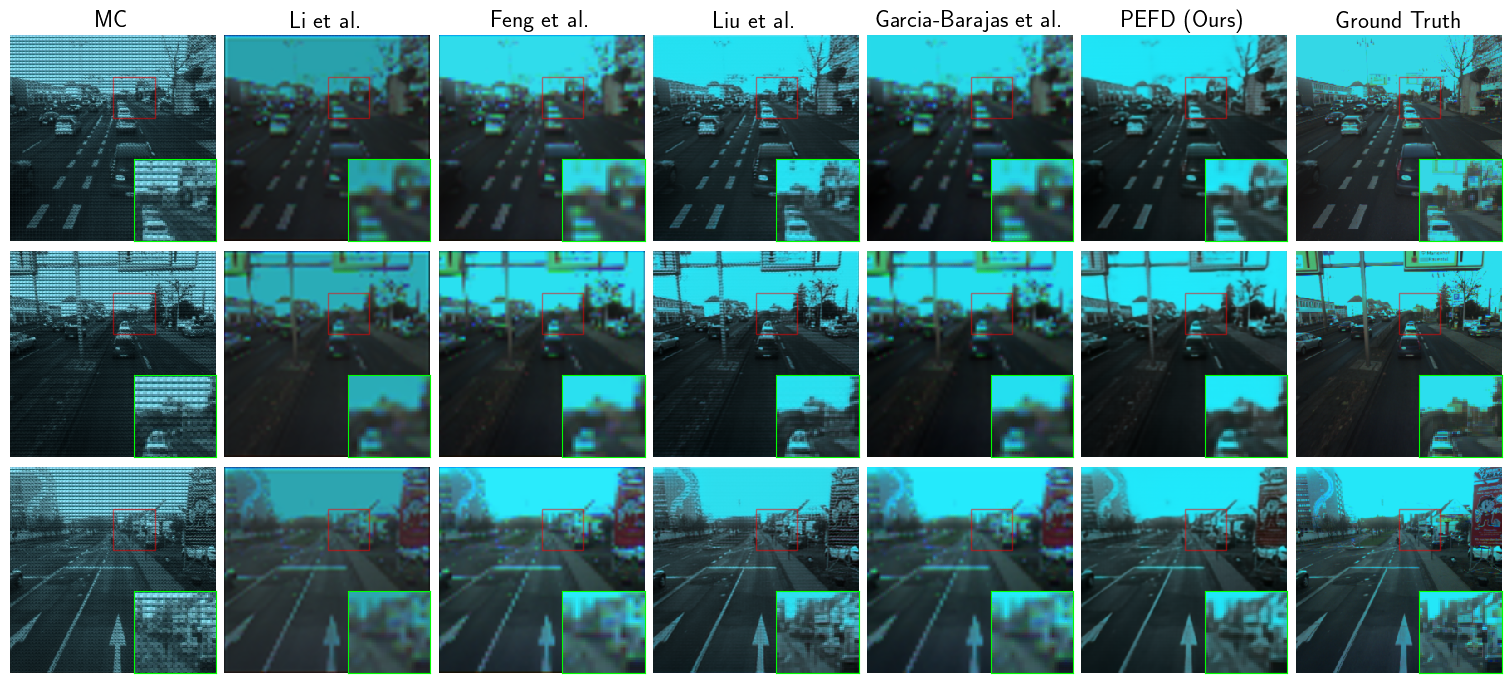}
  \caption{Test set false-RGB reconstruction results on 3 example urban-driving images from the HyKo dataset \cite{winkens_hyko_2017}, showing mosaiced input $\y$, reconstruction results for 9 comparison methods, our reconstruction with PEFD, and ground truth for reference.}
  \label{fig:results_hyko}
\end{figure*}

\begin{figure}[tb]
  \centering
  \includegraphics[width=0.5\textwidth]{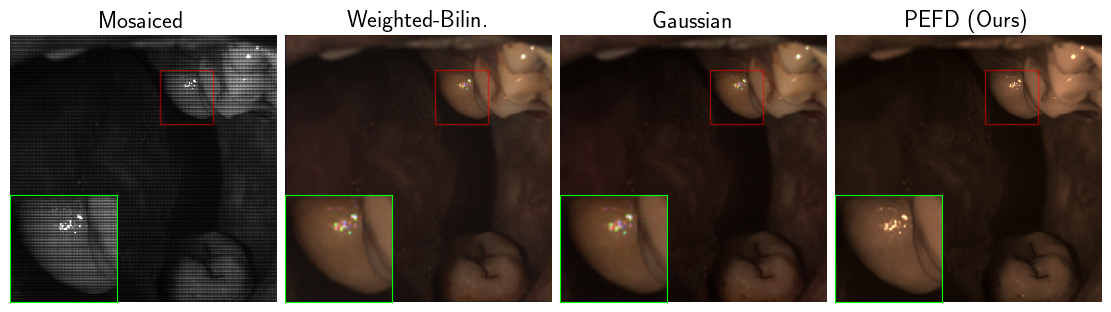}
  \caption{Test-set false-RGB results of demosaicing raw imec CMV2K-SSM4x4-VIS sensor data captured during oral imaging \cite{chand_-vivo_2024}, centre-cropped to a $768\times768$ ROI, showing sensor data $\y$, interpolation and our method. Note no GT is available.}
  \label{fig:results_modid}
\end{figure}

\subsubsection{Evaluation}
We report PSNR and SSIM as standard distortion metrics, capturing pixel-level and structural image fidelity respectively. We additionally report SAM \citep{meng_large-scale_2021} and ERGAS \citep{vivone_critical_2015} as established, complementary multispectral metrics, where SAM captures spectral reconstruction fidelity and ERGAS overall spatial and spectral quality across bands.

\subsubsection{Baselines}
We compare PEFD against several classical interpolation methods: bilinear interpolation with a $5\times5$ kernel, weighted-bilinear \cite{brauers_color_2006} with a $7\times7$ kernel, Gaussian interpolation with a $9\times9$ kernel, and PPID \cite{mihoubi_multispectral_2017}. We also evaluate total variation (TV) with proximal gradient descent and Deep Image Prior (DIP) \citep{ulyanov_deep_2018,park_joint_2020}, using a 16-channel convolutional decoder network \citep{darestani_accelerated_2021}. Finally, we compare against recent self-supervised demosaicing methods: \citet{feng_unsupervised_2024} trains SDNet with shift-EI loss \citep{chen_equivariant_2021}, \citet{garcia-barajas_self-supervised_2025} trains DnCNN \citep{zhang_beyond_2017} with rotate-EI loss \citep{chen_robust_2022}, \citet{li_spatial_2023} trains EDSR \citep{lim_enhanced_2017} with classical regularisation terms, fine-tuning with measurement consistency (MC) \citep{darestani_test-time_2022}, and fine-tuning with limited-pixel shift-equivariant loss from \citet{liu_fine-tuning_2024}.

\subsection{Results}

\begin{table*}[tb]
\centering
\caption{Quantitative results on the HELICoiD \cite{fabelo_-vivo_2019} and HyKo \cite{winkens_hyko_2017} datasets. Best results in \textbf{bold}.}
\label{tab:results}
\begin{subtable}[t]{0.48\linewidth}
\centering
\caption{HELICoiD dataset.}
\label{tab:results_helicoid}
\begin{tabular}{lcccc}
\toprule
Method & PSNR$\uparrow$ & SSIM$\uparrow$ & SAM$\downarrow$ & ERGAS$\downarrow$ \\
\midrule
Mosaiced & 24.09 & 0.330 & 1.366 & 30.70 \\
Bilinear & 40.01 & 0.980 & 0.051 & 6.72 \\
Weighted-Bilin. & 40.62 & 0.982 & 0.038 & 6.17 \\
PPID \citep{mihoubi_multispectral_2017} & 40.77 & 0.979 & 0.038 & 6.90 \\
Gaussian & 40.98 & 0.983 & \textbf{0.034} & 6.13 \\
TV & 21.82 & 0.7705 & 0.091 & 5.84 \\
DIP & 32.15 & 0.914 & 0.138 & 18.69 \\
MC & 35.50 & 0.886 & 0.139 & 13.50 \\
\citet{li_spatial_2023} & 40.65 & 0.970 & 0.039 & 6.18  \\
\citet{feng_unsupervised_2024} & 40.64 & 0.982 & 0.038 & 6.16 \\
\citet{liu_fine-tuning_2024} & 33.11 & 0.901 & 0.343 & 17.20 \\
\small\citet{garcia-barajas_self-supervised_2025} & 40.98 & 0.983 & 0.034 & 6.14 \\
PEFD (Ours) & \textbf{44.84} & \textbf{0.992} & 0.042 & \textbf{4.46} \\
\bottomrule
\end{tabular}
\end{subtable}
\hfill
\begin{subtable}[t]{0.48\linewidth}
\centering
\caption{HyKo dataset.}
\label{tab:results_hyko}
\begin{tabular}{lccc}
\toprule
Method & PSNR$\uparrow$ & SSIM$\uparrow$ & SAM$\downarrow$ \\
\midrule
Mosaiced & 11.33 & 0.072 & 1.330 \\
Bilinear & 30.55 & 0.900 & 0.064 \\
Weighted-Bilin. & 30.38 & 0.896 & 0.055 \\
PPID \citep{mihoubi_multispectral_2017} & 31.46 & 0.901 & 0.052 \\
Gaussian & 32.57 & 0.915 & \textbf{0.044} \\
TV & 23.08 & 0.744 & 0.045 \\
DIP & 28.43 & 0.888 & 0.168 \\
MC & 21.74 & 0.556 & 0.169 \\
\citet{li_spatial_2023} & 29.15 & 0.883 & 0.059 \\
\citet{feng_unsupervised_2024} & 30.95 & 0.900 & 0.052 \\
\citet{liu_fine-tuning_2024} & 30.56 & 0.858 & 0.118 \\
\citep{garcia-barajas_self-supervised_2025} & 32.57 & 0.915 & 0.045 \\
PEFD (Ours) & \textbf{34.81} & \textbf{0.938} & 0.064 \\
\bottomrule
\end{tabular}
\end{subtable}
\end{table*}

\subsubsection{HELICoiD dataset}
Results are shown in \cref{tab:results_helicoid,fig:results_helicoid}. Classical interpolation methods (bilinear, weighted-bilinear, Gaussian, PPID) achieve similar quantiative performance, recovering some spectral information but producing blurred reconstructions that lack high-frequency details (particularly in the fine blood vessels), with performance increasing with kernel size at the expense of sharpness. The Gaussian interpolation successfully removes staircasing artifacts in the demosaiced images, but retains heavy blurring. TV performs poorly, producing heavily cartoonised images due to the extreme sparsity of the $16\times$ undersampling. Among self-supervised methods, \citet{feng_unsupervised_2024,garcia-barajas_self-supervised_2025,li_spatial_2023} and DIP achive higher fidelity compared to classical methods, but still lose fine details in the blood vessels, suggesting limited ability to recover information from the mosaicing null-space. This is due to, respectively: the limited group structure of the equivariant imaging transforms combined with model training from scratch, the limited regularisation effect of classical TV/Tikhonov terms,{\parfillskip=0pt\par}


and the limited intrinsic inductive bias of the network in DIP. The poor performance of fine-tuning with only MC demonstrates that it cannot fine-tune within the null-space of the forward operator. \citet{liu_fine-tuning_2024} recovers good details, but the spectral content is shifted in the reconstructions (see \cref{fig:spectral_examples}), due to the limited size of the within-filter pixel-shift group $\textbf{C}_C$, which is much smaller than the full pixel-shift group $\textbf{C}_W$. PEFD substantially outperforms the baselines in PSNR (almost 4 dB over next best), SSIM and ERGAS by leveraging the larger group structure of perspective-equivariance, and by fine-tuning from a robust pretrained model. We observe that the SAM is higher than that of classical methods. While this may reflect some spectral inaccuracy, we observe that the method's sample spectra shown in \cref{fig:spectral_examples} match the GT well. Qualitatively, PEFD reconstructions preserve fine anatomical structures in the brain and exhibit good spectral consistency with spectral content close to GT.

\subsubsection{HyKo dataset}
Results are shown in \cref{tab:results_hyko,fig:results_hyko}. The ranking of methods is consistent with HELICoiD. Classical interpolation methods contain similar blurring artifacts. In addition, \citet{feng_unsupervised_2024,garcia-barajas_self-supervised_2025,li_spatial_2023} and DIP achieve good reconstructions but with some remaining blur, due to the limited ability of these methods to recover information from the null-space of the mosaicing. Qualitatively, only PEFD recovers sharp edges in the driving scenes such as diagonal line markings, while retaining texture within objects such as other cars and removing gridding artifacts caused by lack of spectral fidelity.

\subsubsection{Raw data experiment}

Qualitative results of demosaicing raw unprocessed sensor data without ground truth are shown in \cref{fig:results_modid}. Interpolation methods recover full-resolution multispectral images, but only at moderate spatial frequency, while generating chromatic artifacts. Conversely, our DL method recovers high-frequency details without chromatic artifacts, and without training on GT images.

\subsection{Ablation analysis}
\label{sec:ablation}

We ablate the loss function of PEFD by comparing with RAM zero-shot (foundation model without fine-tuning), RAM fine-tuned with shift-equivariance \citep{chen_equivariant_2021} (\ie \citet{feng_unsupervised_2024} but no longer from scratch) and supervised oracle fine-tuning as a gold-standard. Results are shown in the supplementary \cref{sec:ablation_supp}. RAM zero-shot performs poorly as expected, since the 16-channel weights were initialised from grayscale data and thus cannot capture cross-spectrum correlations. Fine-tuning with shift-equivariance improves performance substantially, but exhibits mosaic artifacts due to the limited spatial transformation set. These results demonstrate that perspective-equivariance enables effective self-supervised learning without GT, approaching supervised performance.

\section{Conclusion}
\label{sec:conclusion}
We propose PEFD, a framework that learns multispectral demosaicing from mosaiced measurements alone. PEFD exploits the projective geometry of snapshot camera-based imaging systems to leverage a richer group structure than previous demosaicing methods, recovering more null-space information without ground truth (GT). By adapting pretrained foundation models designed for 1-3 channel imaging, PEFD learns efficiently without the costly GT obtained from slow line-scanning systems. On intraoperative and automotive datasets, PEFD recovers fine details such as blood vessels and preserves spectral fidelity, substantially outperforming recent approaches and nearing supervised performance. Furthermore PEFD is validated on a raw, unprocessed dataset where GT truly does not exist.

Realistically, deployed systems will require a mix of both supervised post-training on adjacent datasets and self-supervised tuning as we have demonstrated here; future work would explore engineering techniques to maximise performance. Further applications include adding temporal symmetries for video demosaicing, as well as other forms of compressive snapshot imaging such as CASSI systems.
\clearpage
{
    \small
    \bibliographystyle{ieeenat_fullname}
    \bibliography{references}
}

\clearpage
\setcounter{page}{1}
\maketitlesupplementary

\section{Ablation analysis results}
\label{sec:ablation_supp}

\begin{figure*}[b]
  \centering
  \begin{subfigure}[t]{0.4\textwidth}
    \includegraphics[width=\textwidth]{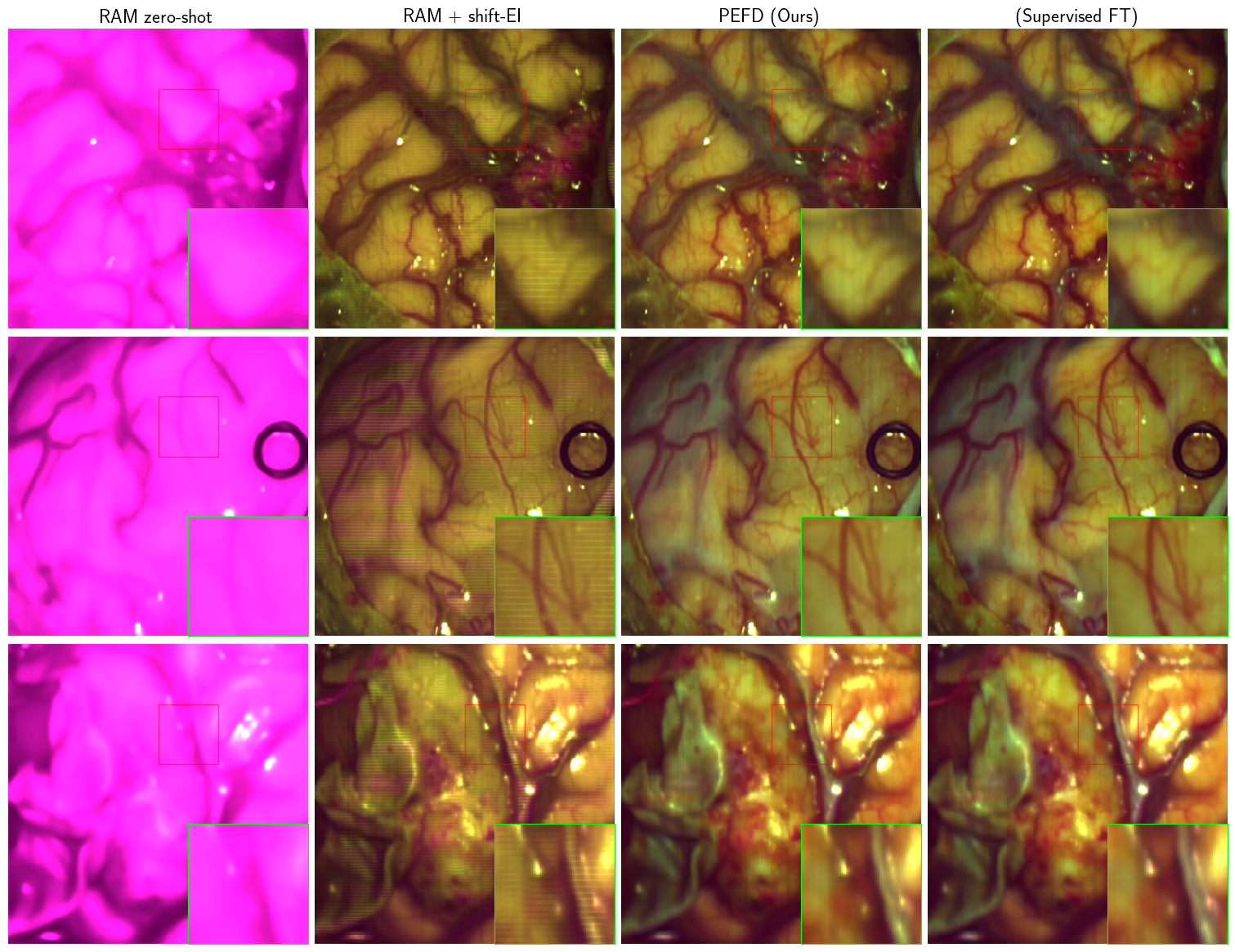}
  \end{subfigure}%
  \begin{subfigure}[t]{0.4\textwidth}
    \includegraphics[width=\textwidth]{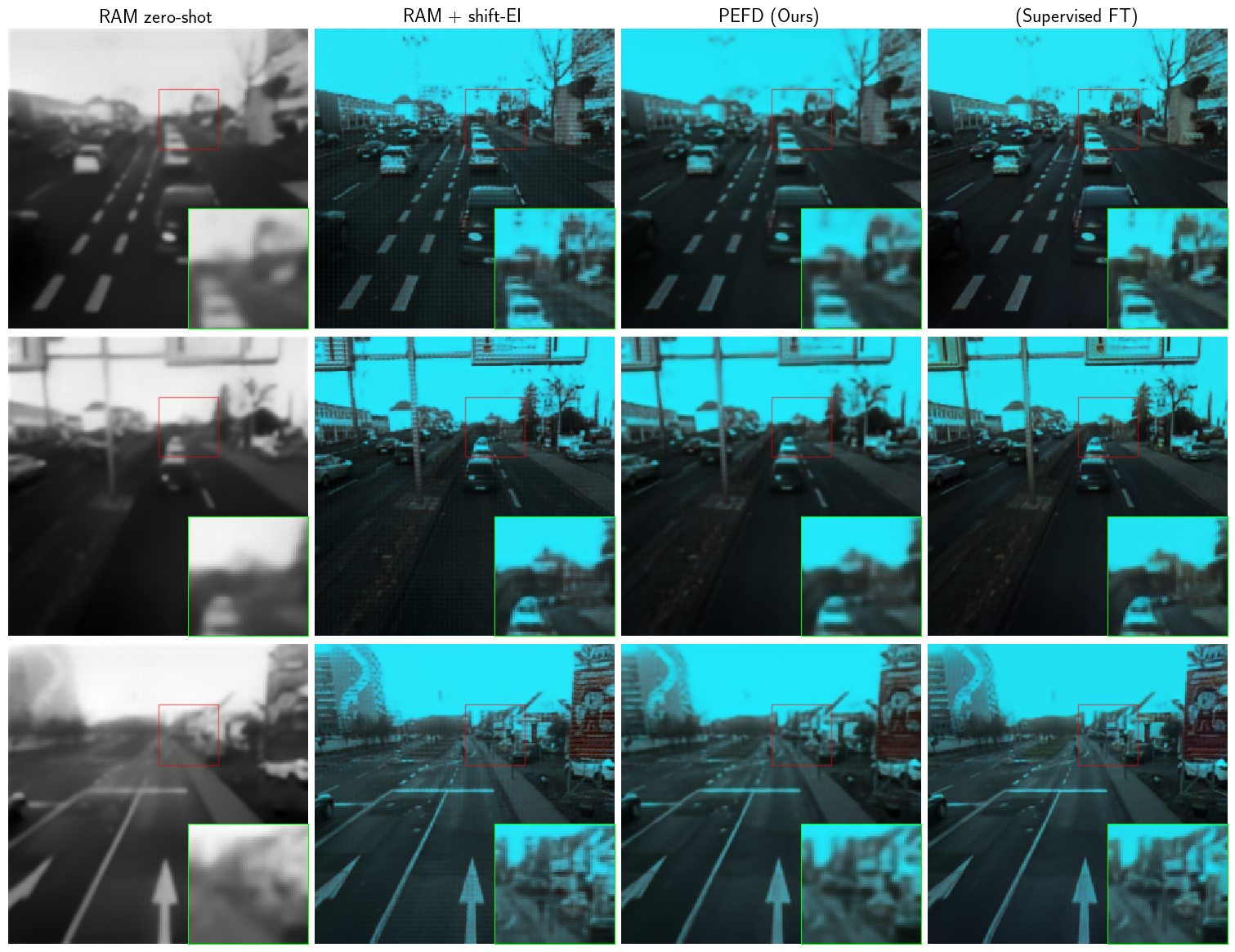}
  \end{subfigure}
  \caption{Ablation study on 3 example neurosurgical images from HELICoiD \cite{fabelo_-vivo_2019} (left) and 3 example urban-driving images from the HyKo dataset \cite{winkens_hyko_2017} (right).}
  \label{fig:results_ablation}
\end{figure*}

We present the quantitative and qualitative results of the ablation analysis in \cref{tab:ablation,fig:results_ablation}. We refer the reader to the main paper \cref{sec:ablation} for discussion.

\begin{table*}[b]
\centering
\caption{Ablation study on the HELICoiD \cite{fabelo_-vivo_2019} and HyKo \cite{winkens_hyko_2017} datasets. Best self-supervised result in \textbf{bold}.}
\label{tab:ablation}
\begin{subtable}[t]{0.48\linewidth}
\centering
\caption{HELICoiD dataset.}
\label{tab:ablation_helicoid}
\begin{tabular}{lcccc}
\toprule
Method & PSNR$\uparrow$ & SSIM$\uparrow$ & SAM$\downarrow$ & ERGAS$\downarrow$ \\
\midrule
RAM zero-shot \citep{terris_reconstruct_2025} & 28.00 & 0.759 & 0.631 & 50.10 \\
RAM + shift-EI \citep{chen_equivariant_2021} & 41.77 & 0.979 & 0.087 & 6.76 \\
PEFD (Ours) & \textbf{44.84} & \textbf{0.992} & \textbf{0.042} & \textbf{4.46} \\
\midrule
(Supervised FT) & 46.83 & 0.995 & 0.028 & 3.60 \\
\bottomrule
\end{tabular}
\end{subtable}
\hfill
\begin{subtable}[t]{0.48\linewidth}
\centering
\caption{HyKo dataset.}
\label{tab:ablation_hyko}
\begin{tabular}{lccc}
\toprule
Method & PSNR$\uparrow$ & SSIM$\uparrow$ & SAM$\downarrow$ \\
\midrule
RAM zero-shot \citep{terris_reconstruct_2025} & 18.59 & 0.807 & 0.318 \\
RAM + shift-EI \citep{chen_equivariant_2021} & 34.47 & 0.909 & 0.069 \\
PEFD (Ours) & \textbf{34.81} & \textbf{0.938} & \textbf{0.064} \\
\midrule
(Supervised FT) & 38.18 & 0.968 & 0.032 \\
\bottomrule
\end{tabular}
\end{subtable}
\end{table*}

\end{document}